\documentclass[lettersize,journal]{IEEEtran}

\usepackage{amssymb} 
\usepackage[T1]{fontenc}
\usepackage[utf8]{inputenc} 

\usepackage{graphicx}
\usepackage{array}

\usepackage{listings}
\usepackage[most]{tcolorbox}

\usetikzlibrary{arrows.meta} 
\usepackage{tikz}
\usetikzlibrary{positioning, arrows.meta, fit, backgrounds}
\usetikzlibrary{arrows.meta, positioning, shapes.multipart}

\usepackage[]{subcaption}
\usepackage{caption}
\usepackage{xcolor}

\definecolor{mfok}{RGB}{0,128,0}
\definecolor{mftodo}{RGB}{128,0,0}

\usepackage{enumitem}

\usepackage{url}

\usepackage{titlesec}

\newcommand{\paragraf}[2]{
	\par \vspace{0.5ex}\noindent\textbf{#1:} #2
}

\newcommand{\paragrafSimple}[2]{
	\par \vspace{1ex}\noindent\emph{#1:} #2
}

\usepackage{pifont}
\newcommand{\cmark}{\ding{51}} 
\newcommand{\xmark}{\ding{55}} 

\usepackage[table]{xcolor} 
\definecolor{lightblue}{RGB}{173, 216, 230}
\definecolor{khaki}{RGB}{240, 230, 140}
\definecolor{lightgreen}{RGB}{144, 238, 144}
\usepackage[table,xcdraw]{xcolor}  %
\usepackage{amssymb}

\newcommand{\mfb}[1]{\textcolor{black}{#1}}
\newcommand{\rslMathSymbol}[1]{\ensuremath{#1}}
\newcommand{\RSTM}{\rslMathSymbol{\rm RS(TM)^2}}

\begin{document}

\setlist[itemize]{leftmargin=0.35cm} 

\setlist[enumerate]{leftmargin=0.5cm}

	\title{Ontology-Driven Robotic Specification Synthesis}
	
	\author{Maksym Figat$^{a,b}$,~\IEEEmembership{Member,~IEEE}, 
     Ryan M. Mackey$^{b}$, 
    Michel D. Ingham$^{b}$%
	\thanks{$^{a}$Maksym Figat (\emph{corresponding author}) is with Warsaw University of Technology (WUT), Warsaw, Poland. \texttt{maksym.figat@pw.edu.pl}}%
	\thanks{$^{b}$The authors are with NASA Jet Propulsion Laboratory (JPL), California Institute of Technology, Pasadena, CA, USA. Maksym Figat was a Visiting Postdoctoral Researcher at JPL under the Bekker Fellowship of the Polish National Agency for Academic Exchange (NAWA) during this work. This research was conducted in part at the Jet Propulsion Laboratory, California Institute of Technology, under a contract with the National Aeronautics and Space Administration (80NM0018D0004), and was partially supported by the Excellence Initiative: Research University (IDUB) program at WUT.}
	\thanks{{\scriptsize This work has been submitted to the IEEE for possible publication. Copyright may be transferred without notice, after which this version may no longer be accessible.}}
	}


\maketitle

\begin{abstract}

\mfb{
This paper addresses robotic system engineering for safety- and mission-critical applications by bridging the gap between high-level objectives and formal, executable specifications. The proposed method, \emph{Robotic System Task to Model Transformation Methodology} (\RSTM{}) is an ontology-driven, hierarchical approach using stochastic timed Petri nets with resources, enabling Monte Carlo simulations at mission, system, and subsystem levels. A hypothetical case study demonstrates how the \RSTM{} method supports architectural trades, resource allocation, and performance analysis under uncertainty.  Ontological concepts further enable explainable AI-based assistants, facilitating fully autonomous specification synthesis. The methodology offers particular benefits to complex multi-robot systems, such as the NASA CADRE mission, representing decentralized, resource-aware, and adaptive autonomous systems of the future.
}
\end{abstract}


\section{Introduction}
\mfb{Development of robotic systems for dynamic, high-uncertainty environments is complicated by limitations of environmental modeling, interactions between robots and humans, and unforeseen or emergent behaviors.  In order to ensure correct decision-making and task execution that meets reliability and safety standards, we must address several challenges in software engineering for robotics, as described in~\cite{Goues:2024}:  Layered, leaky system heterogeneity, integration of machine learning, and siloed development.  Solution requires multi-view hierarchical architecture description languages supporting multiple design views and analyses, and stresses the need for a reference robotics architecture, as explained in~\cite{Brugali:2024}.  Current tools lag behind technological advances, making it impossible to produce reliable control code from requirements without considerable input from the designer, and without encountering gaps between simulation and implementation.}

\mfb{The problem may be greatly reduced by adding an intermediate step between requirements and implementation, that of system specification, and producing corresponding models.  Such an approach is consistent with Model-Based System Engineering (MBSE), proceeding from a model, a meta-model, and a domain-specific language to describe the system from multiple viewpoints.  Model-Driven Engineering (MDE) treats the model as an artifact for automatic model-to-model and model-to-code transformation, though MDE is still hampered by the lack of a holistic abstract view of the robotic system~\cite{nordmann:2016:new,Silva:2021}.  Models may be informal (e.g., SysML~\cite{Friedenthal:2015,Wagner:2012}) or formal (e.g., Finite State Machine (FSM) or Petri nets (PN)~\cite{Figat:2022:RAS,Pelletier:2025:RAS}) with the latter enabling formal property analysis.}

\mfb{Combining MBSE with formal models holds promise, especially for formal models spanning multiple abstraction layers.  An example is found in Robotic System Specification Methodology (RSSM)~\cite{Figat:2022:RAS} based on a parametrized Robotic System Hierarchical PN metamodel (RSHPN); the RSSM parameters are in turn defined in the Robotic System Specification Language (RSSL), enabling automatic generation of ROS controller code~\cite{Figat:2022:RAL}.}

\mfb{Current approaches assume that the designer manually develops a system model, integrating information about the environment, resources, devices, and constraints from multiple system perspectives, a gap that motivated the \RSTM{} approach.  \RSTM{} is based on a robotic system ontology inspired by IEEE 1872.2-2021~\cite{ieee-standard:2021:autonomous-robotics}, defining interactions between a robotic system and its environment; and IEEE 1872.1-2024~\cite{ieee-standard:2024:task}, defining core task-related concepts such as \emph{action}, \emph{plan}, \emph{task}, \emph{constraint}, and \emph{resource}. Related work~\cite{Pinto:2021} also addresses system-environment interactions and capabilities in autonomous systems.}

\mfb{Our ontology integrates these approaches but is limited to the detail necessary to model at three levels of abstraction -- Mission, System, and Subsystem -- providing a hierarchical view of the specification, extending it with system capabilities, object affordances, and resources required for each entity to perform an action.  The three levels capture entity behavior and environment interaction using stochastic, timed, resource-aware PNs, validated through Monte Carlo simulation. The final, subsystem-level model produces parameters for the RSSM, enabling synthesis of specifications and subsequent automatic synthesis of ROS 2 controller code.}

\mfb{The \RSTM{} approach offers three main advantages:  (1) A bridge across the specification gap from general objectives to formal, multi-level specifications via robotics-specific ontology; (2) Executable specifications, producing stochastic, timed, resource-aware PNs suitable for Monte Carlo simulation; and (3) Automatic synthesis of ROS 2 code from validated specifications.  We illustrate the process with an example, that of a hypothetical robot team designed to locate and manipulate objects, building a vertical stack.}


\section{\RSTM{} Overview}
\label{sec:rstm}
\begin{figure}
	\centering
	\includegraphics[width=0.9\linewidth]{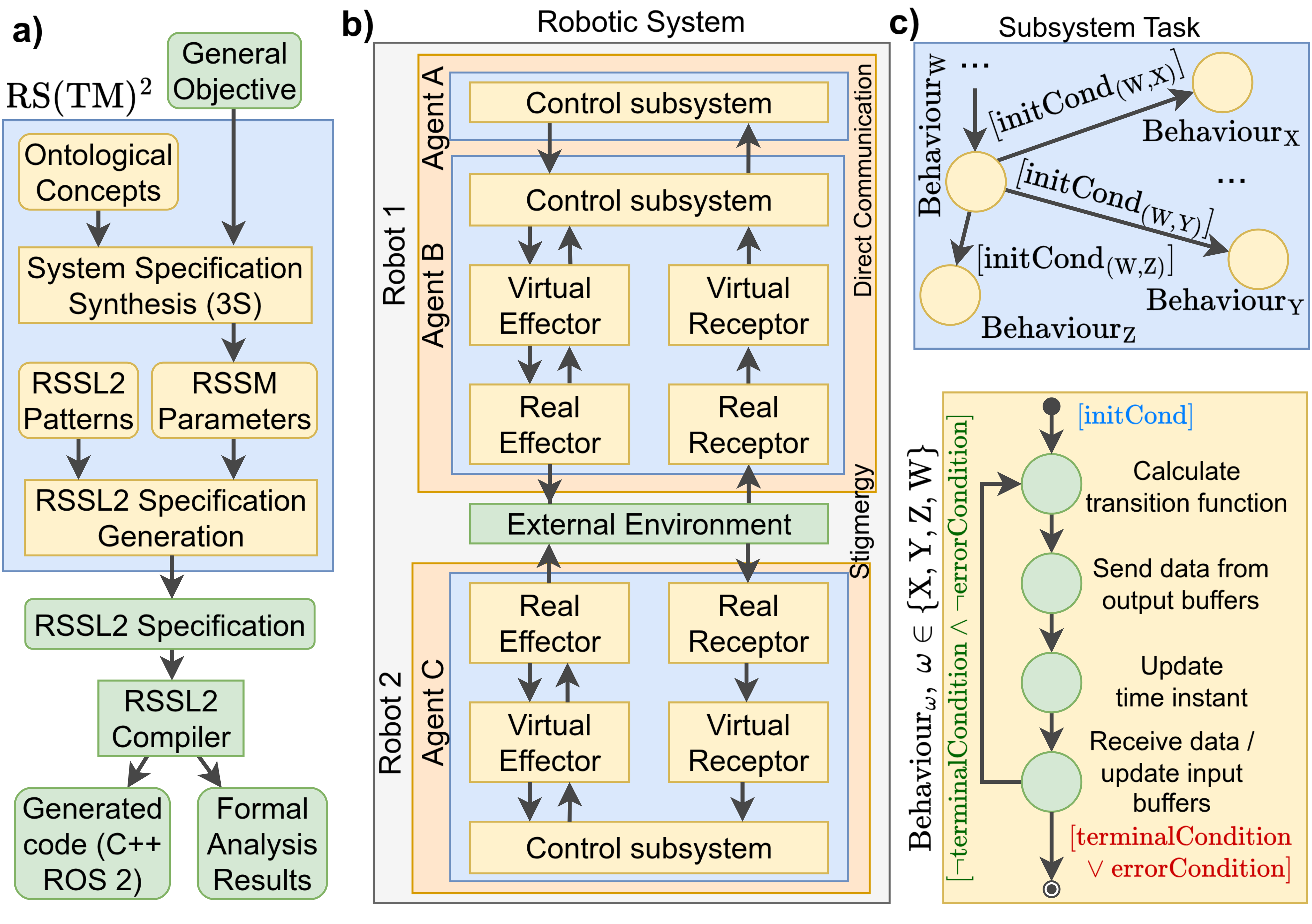}
	\caption{(a) \RSTM{} procedure; (b) basic RSSM structural concepts; (c) basic RSSM activity concepts.}
	\label{fig:rstm_rssm}
\end{figure}

\mfb{\RSTM{} focuses on the specification gap, i.e., the gap between requirements and implementation, defined as follows: Given a \emph{General Objective} as input, it produces a synthesized specification (expressed in \emph{RSSL2} specification language) as output. The process consists of two procedures, as described in Fig.~\ref{fig:rstm_rssm}a:
\begin{enumerate}
	\item \emph{System Specification Synthesis (3S)}, deriving the system model parameters,
	\item \emph{RSSL2 Specification Generation}, using specification language patterns to transform these parameters into the specification representing the system model.
\end{enumerate}
A key feature of the method is its use of ontological concepts, providing a common semantic foundation across all specification levels for 3S design steps, described below.}

\paragraf{System Specification Synthesis}{\mfb{The 3S procedure yields a set of system model parameters defining the \emph{structure} and \emph{activity} of the robotic system (RS), grounded in formal concepts in~\cite{Figat:2022:RAS}.}}

\textit{Structure:} \mfb{We define an RSSM-based robotic system to be composed of robots and auxiliary devices.  Each robot comprises one or more embodied agents (Fig.~\ref{fig:rstm_rssm}b).  Each agent includes a control subsystem, coordinating receptors and effectors, both real (physical devices interfacing with the environment) and virtual (internal representations supporting abstraction and preprocessing).  Subsystems include internal memory, input buffers, and output buffers.  Communication occurs via inter-agent channels (viz., between control subsystems) as well as intra-agent channels (between subsystems within an agent).}

\textit{Activity:} \mfb{Each subsystem operates by executing behaviors at a predefined frequency (Fig.~\ref{fig:rstm_rssm}c). A behavior is selected according to initial condition (i.e., persistent state, goals, or commands) and runs in an iterative loop. In each iteration, at the subsystem's predefined frequency, it computes a transition function that produces data based on internal memory and input data from other subsystems. The subsystem then sends data from output buffers to other subsystems, updates its input buffers.  It also checks terminal and error conditions, and if met, the behavior terminates and the next behavior executes according to updated initial conditions.}

\paragraf{RSSL2 Specification Generation}{
\mfb{Following the procedure above, the system model parameters defining the robotic system are formalized into the specification using a dedicated Visual Studio Code plug-in, providing parametrized templates.  This significantly reduces dependency on knowledge of the language syntax.  A custom RSSL2 compiler, developed as part of this effort, transforms specifications into distributed C++ code.  Each subsystem is mapped to a separate ROS~2 node, with communication handled via topics.}}
 

\section{\RSTM{} Ontological Concepts}
\label{sec:ontology}
\mfb{A coherent set of ontological concepts is required to systematically derive the robotic system model from high-level task objectives, structuring system architecture, behavior, and environmental interaction.  Each concept may be instantiated at different levels (mission, system, or subsystem) depending on the specification phase: Vocabulary remains consistent, but role and granularity of each concept adapts to the level.  A core subset of concepts playing a central role is explained below:}

\begin{figure}
	\centering
	\includegraphics[width=1.0\linewidth]{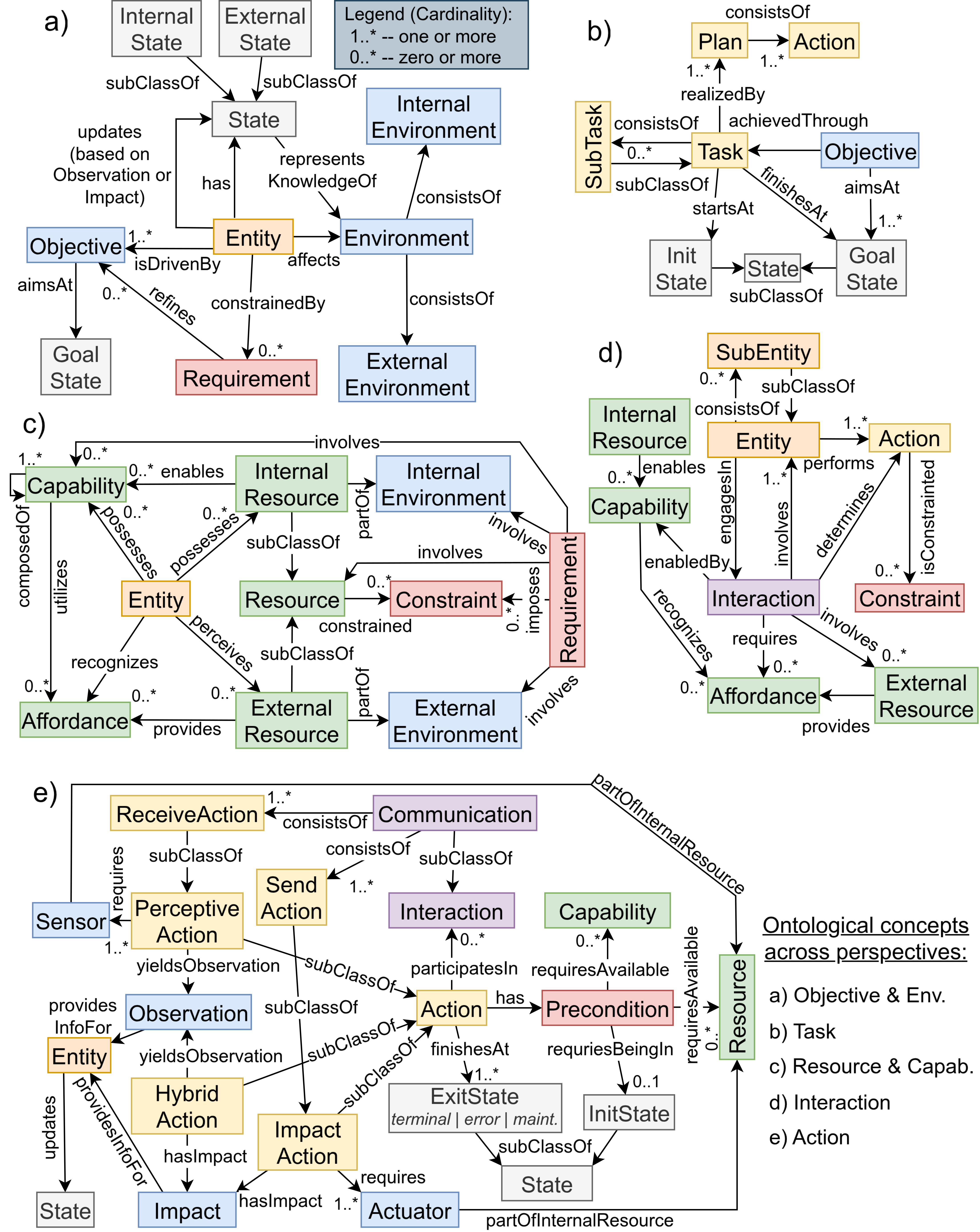}
	\caption{Ontological concepts from different perspectives}
	\label{fig:all-ontological-perspectives}
\end{figure}

\paragraf{Entity, Objective and Environment}{ 
\mfb{See diagram in Fig.~\ref{fig:all-ontological-perspectives}a.  Specification begins by selecting the \emph{entity}, i.e., the subject of design at a chosen level, followed by general description of its objective. The designer then formulates \emph{requirements} to achieve that \emph{objective}.  It is essential to describe the entity’s operating \emph{environment}, distinguishing between external and internal environment. Entity knowledge about the environment is represented as \emph{state}. State evolves over time based on \emph{observations} made by the entity and outcomes of its \emph{actions} (described as \emph{impact}). Only when environment, state, and requirements have been defined can the entity’s objective be refined, performed by specifying one or more goal states that the entity attempts to achieve.}}

\paragraf{Task}{\mfb{Described in Fig.~\ref{fig:all-ontological-perspectives}b. Each objective is associated with a specific goal state. Achieving this state involves a \emph{task} which transforms \emph{initial state} of the environment into the goal state.  The task may be decomposed into \emph{subtasks} with intermediate goal states.  The task is realized through execution of a \emph{plan}. Multiple realizations may exist, i.e., different plans with different sequences of actions.}}

\paragraf{Resource and Capability}{\mfb{See Fig.~\ref{fig:all-ontological-perspectives}c. Realization of an objective requires suitable \emph{capabilities} and access to resources.  The external environment may provide external resources, or the entity may have internal resources such as sensors, actuators, or components providing computation, memory, or energy. Internal resources enable basic capabilities needed for \emph{interaction}.  Objects in the environment offer \emph{affordances}~\cite{Gibson:1979}, i.e., potential means of interaction, but only affordances aligned with the entity’s capabilities can be recognized and utilized.  The environment is therefore perceived not only through physical properties but through the entity’s capabilities, e.g., a gripper capability allows an entity to perceive the graspable affordance of a block and thereby interact with it.  Interaction is governed by affordance-capability pairs, and composed capabilities may emerge from combinations of basic capabilities.}}

\paragraf{Interaction}{\mfb{As shown in Fig.~\ref{fig:all-ontological-perspectives}d, potential interactions between the entity and its surroundings can be identified once the entity’s capabilities and affordances of objects in the environment are defined. An interaction can occur when an object offers a relevant affordance, and the entity can exploit it through a corresponding capability. Each interaction describes a potential action the entity can perform with respect to the environment, subject to \emph{constraints} such as resource usage or execution time.  We may decompose an entity into sub-entities, each corresponding to a specific task, if multiple interactions must take place concurrently.  Additionally, some interactions may require multiple entities (e.g., a team carry action), in which case each individual entity views the external entities as providing a relevant affordance.  Note that multiple coordinated local interactions may be interpreted as a higher-level activity conducted by a collective entity; if there is no explicit higher-level coordination, this describes emergent behavior.}}

\paragraf{Action}{\mfb{Described in Fig.~\ref{fig:all-ontological-perspectives}e, an action results from an interaction that combines the entity’s capability with an environmental affordance.  The action may involve affecting the environment (\emph{impact action}) or observing it (\emph{perceptive action}).  Impact actions use \emph{actuators} to produce an impact and may be probabilistic. Perceptive actions use \emph{sensors} to gather observations.  Both contribute to changes in entity internal state.  \emph{Hybrid actions} combine impact and perceptive actions, e.g., heating a sample to a specified temperature, which requires acting on the environment as well as sensing its temperature property.  \emph{Communication} is also an action; it may be direct or mediated by the environment, and enables asynchronous exchange.}}

\mfb{Action execution depends on preconditions defined over available capabilities, resources, and environment state.  Actions may have durations and produce impacts.  Execution of an action ends when one of three conditions are met:  \emph{Terminal}, when the desired state is reached; \emph{Error}, when a fault or malfunction occurs; and \emph{Maintenance}, when required resources or capabilities become unavailable.  Terminal and Error conditions trigger action termination, whereas Maintenance conditions affect the ability to continue or require switching to another action due to resource loss.} 

\begin{figure}
\centering
\includegraphics[width=0.95\linewidth]{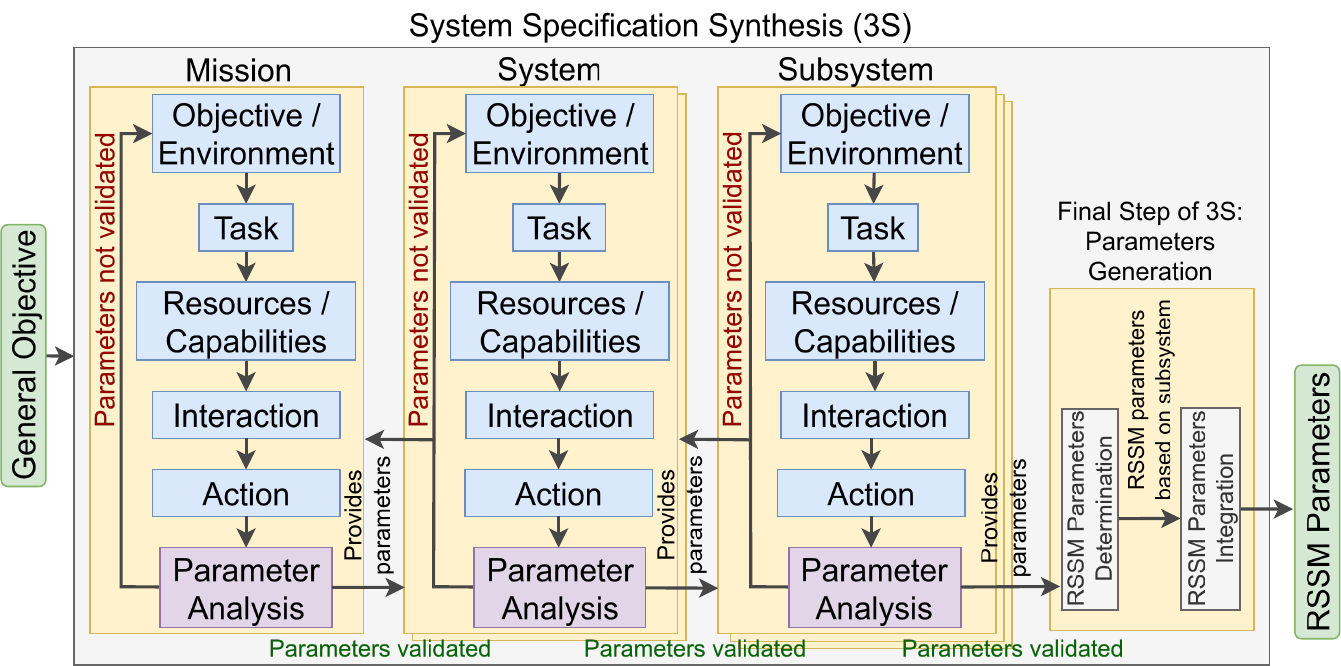}
\caption{Three 3S levels: Mission, System and Subsystem, each with a distinct specification perspective.}
\label{fig:3s-idea-phases}
\end{figure}

\section{System Specification Synthesis (3S)}
\label{sec:3s}

\mfb{We will describe the 3S procedure for a classic three-level design, viz., mission, system, and subsystem, though the technique can be adapted for additional levels if desired.  Each level offers a distinct view of the robotic system and corresponds to a specific abstraction defined in the ontological concepts.  This layered view introduces a horizontal perspective in which each entity (mission, system, or subsystem) adds increasing detail.  In parallel, the vertical perspective (Fig.~\ref{fig:3s-idea-phases}) shows that all entities are defined using the same ontological concepts, with varying interpretation and granularity across abstraction levels.  At the mission level, the environment encompasses the entire operational domain, objectives reflect global intent, and capabilities describe system-wide functions.  At the system level, the entity integrates capabilities and resources provided by its subsystems, forming higher-level and more abstract capabilities used to reason about interaction with the environment in broader scope. This may take the form of a plan, i.e., a structured set of actions leading to the objective, defined with respect to available capabilities and resources.}

\mfb{At the subsystem level, or at the lowest level of a multiple-tier description, definitions become concrete and technically detailed.  Objectives are linked with specific sensors, actuators, and actions.  This layered refinement allows structured decomposition and coherent integration.  While subsystems may operate in restricted environments with limited actions, detailed specification may still be required due to direct interaction with real devices and their constraints.}

\mfb{Refinement across the different levels is iterative: Concepts defined at one level provide the basis for identification of sub-entities at the next.  Flawed decisions may propagate, but an executable specification enables early detection of errors and adjustment of parameters.}


\begin{figure}
	\centering
	\includegraphics[width=1.0\linewidth]{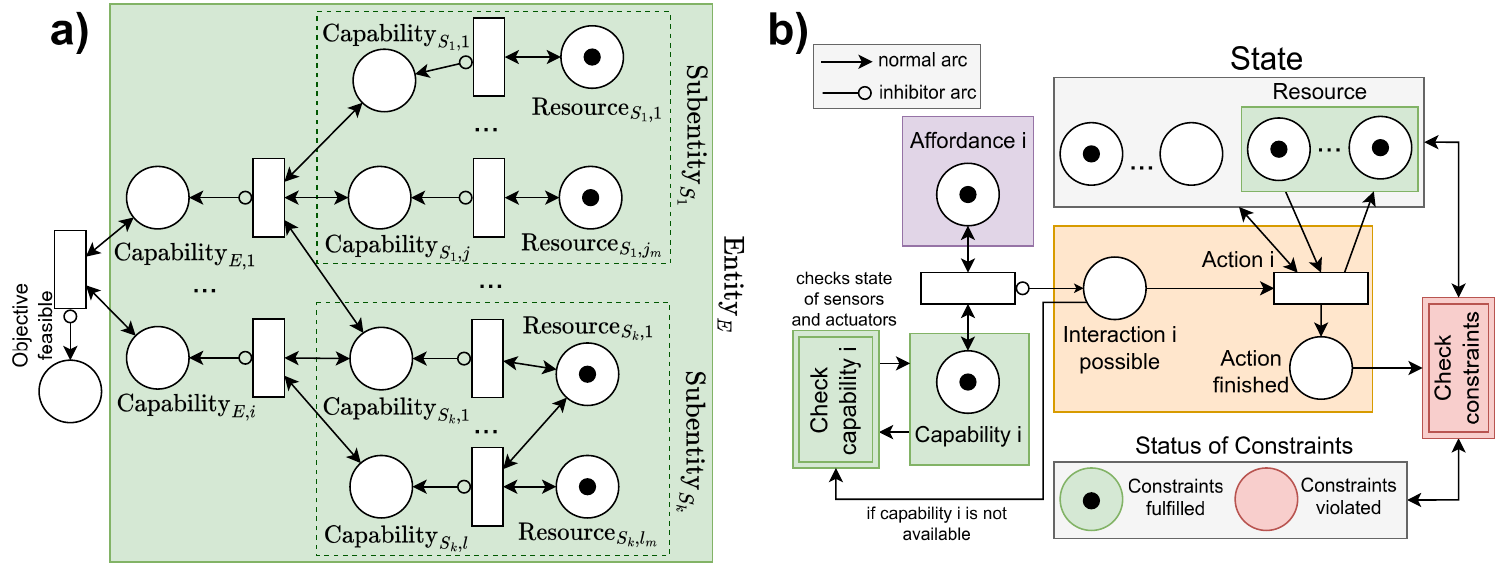}
	\caption{(a) Merged PNs at entity level; Standard arcs (arrows) transfer tokens; inhibitor arcs block transitions; (b) Action feasibility: preconditions, resources, and capabilities.}
	\label{fig:pn_analysis_capabilties_action_complexity}
\end{figure}

\paragraf{Formalism and Analysis}{\mfb{At each level, the 3S parameters are embedded in an extended stochastic timed Petri net with resources (PN). This formalism extends the standard Petri net~\cite{Murata:1989:Petri:Nets} to incorporate time, uncertainty, and resource constraints.  It also supports concurrent execution of multiple instances of a given transition. This formalism yields an executable 3S specification enabling simulation-based analysis of behavior and supporting architectural decisions such as decomposition or resource allocation. It supports multiple analyses, including capability availability, potential for parallel action execution, and objective feasibility.}}

\mfb{The Petri net consists of places (circles), transitions (rectangles), and tokens (black dots): Places represent conditions, resources, or internal states, while transitions correspond to timed actions.  The marking (viz., token distribution and resource state, where resources are represented as non-negative real values) captures the entity’s current state, with the initial marking encoding available conditions and resources.}

\mfb{Each transition instance runs independently, initiating its associated action with a duration sampled from a distribution (e.g., normal or uniform) and then tracked to completion. During execution, the transition instance updates resources continuously according to defined rates, and tokens are deposited only once the transition has finished.}

\mfb{A transition may fire only if sufficient tokens are available and resources can be reserved for the full action duration (while respecting defined resource constraints, e.g., energy limits).  Competing enabled transitions can be selected according to different policies:  Fixed priorities, random choice, or probabilities derived from priority values.  Inhibitor arcs~\cite{Murata:1989:Petri:Nets} may block a transition from firing or may suspend an ongoing action if a token appears in the associated place during its execution.}

\mfb{Achieving an objective requires specific capabilities, which in turn depend on internal and external resource availability.  This forms a dependency chain represented as a PN linking objectives, capabilities, and resources. Entity capabilities are provided by subentitites, each described by its own PN and composable into a hierarchical model (Fig.~\ref{fig:pn_analysis_capabilties_action_complexity}a).  This enables reasoning about mission-level capabilities while considering resource availability and constraints at the subsystem level. The model also informs design-time decisions on reliability and redundancy, such as evaluating the benefit of redundant sensors. Beyond capability and resource analysis, the PN formalism also supports design-correctness checks, state reachability, and liveness analysis.}

\mfb{While introducing time and stochasticity complicates exact verification, approximate methods such as Monte Carlo simulation can be applied.  By varying the initial marking, alternative scenarios can be evaluated to determine which capabilities remain available and how the system responds.  At runtime, the same PN can model resource status and trigger secondary objectives when capabilities are lost.}

%
\mfb{At the capability level, analysis typically focuses on physical resources enabling actions (e.g., sensors, actuators).  As shown in Fig.~\ref{fig:pn_analysis_capabilties_action_complexity}b, an action executes only when its precondition is met, captured by a specific PN marking that encodes the entity’s required internal state and resources.  Execution updates the state and may trigger constraint checks, such as resource limits.  However, action feasibility also depends on consumable resource availability.  For example, if two actuators and sufficient consumables are present, two instances of the same transition may run in parallel, and we may wish to introduce sub-entities responsible for concurrent actions.}

\mfb{PNs offer a powerful formalism for specifying, simulating, and verifying entity behavior during design, enabling Monte Carlo methods of assessment for robustness, capability degradation, and resource-dependent behavior.  Although analysis follows the mission-system-subsystem level hierarchy, it remains iterative and refinement-driven.  Constraints from higher levels propagate downward and become more detailed.  Subsystem decomposition in this manner may expose infeasibility due to timing, resource conflicts, or performance limits, prompting changes to constraints, hardware, or safety and performance assumptions.  These refinements result in updates to parameters or architectural changes at higher levels, after which the PN modeling process is repeated until closure is achieved.}

\paragraf{System Model Parameter Generation}{\mfb{The process completes by mapping 3S concepts to parameters in the RSSM system model.  Each level is represented by a dedicated agent:  The Mission Agent governs mission-level planning, System Agents manage system-level entities (typically individual robots), and Subsystem Agents handle execution on physically integrated subsystems (typically functional groups within a robot).}}

\mfb{Physically integrated subsystems refer to sets of interconnected devices, with each set modeled as a single agent with a unified control subsystem.  For each device within such a subsystem, a pair of virtual and real subsystems is defined (e.g., virtual effector $\rightarrow$ real effector, virtual receptor $\leftarrow$ real receptor) where the real subsystem represents the physical device and the virtual one represents the logical interface.  Communication paths yield buffers for structured data exchange, while internal states and resources are mapped to internal memory.  Entity-level plans decompose into RSSM behaviors, expressed as Petri nets and associated with RSSM subsystems.  Each behavior defines a transition function, terminal and error conditions, with preconditions derived from capabilities, resources, and imposed constraints.}

\mfb{This decomposition is one possible transformation from the 3S specification to an RSSM system model, shaped by architectural decisions derived from mission-level objectives and constraints.  Features of the selected architecture (e.g., reactive vs. deliberative, centralized vs. decentralized) are reflected in the number of agents, their responsibilities, and communication patterns.  For example, in a distributed multi-robot architecture, robot coordination may be handled by System Agents, whereas a single System Agent or even the Mission Agent would coordinate robots in a centralized architecture.} 

\section{Case study: Tower Building System}
\label{sec:case-study}
\mfb{We illustrate the process with an example. Consider a mission requirement to develop a robotic system capable of locating green boxes and stacking three of them atop a red box.  We will further introduce resource constraints by requiring task completion within a fixed interval of 10 minutes and to reserve 10\% of system energy.  This problem decomposes as follows:}

\paragrafSimple{A. Parameter definition}

\noindent\mfb{\textbf{Mission Level:}  Flat 10m x 10m experiment area containing three green boxes and one red box.  \emph{Resources}:  Three green boxes.  \emph{Constraints}:  Complete within 10 min; system may not exceed energy limit.  \emph{Goal state}:  Green boxes stacked atop red box. \emph{Affordances}: Boxes stackable. \emph{Capabilities}: Box stacking. \emph{Actions}: Stack box.  \emph{Requirements}:  Objective and constraints satisfied.}

\noindent\mfb{\textbf{System Level:} 
\emph{Resources (external)}:  Three green boxes $B = 3$, a red box (tower base).  
\emph{Resources (internal)}:  One robot ($R = 1$) composes of subsystems, energy $E$.
\emph{Initial State}:  Green boxes randomly placed, robot initialized, energy at maximum.
\emph{Constraints}:  $T \leq T_{\rm max}=10\rm min$; $E \ge 10\%$.
\emph{Affordances}:  Boxes detectable, trackable, graspable, placeable.
\emph{Capabilities}:  Object detection (localize boxes), object tracking (update box position), object approach (move toward target), object manipulation (lift / place object).
\emph{Actions}:  Detect object, track box, approach object, manipulate box (pick up / place).}

\noindent\mfb{\textbf{Subsystem Level:}  Locomotion, Gripper, Mast, and Camera, summarized in Table~\ref{tab:usecase_subsystems}.}

\begin{table}[t]
	\centering
	\caption{Subsystem specifications}
	\scriptsize
	\renewcommand{\arraystretch}{1.05}
	\begin{tabular}{|p{1.2cm}|p{6.1cm}|}
		\hline
		\textbf{Subsystem} & \textbf{Parameters} \\
		\hline
		Locomotion & \textit{Resources:} Motors, encoders, IMU, energy $E_l$; 
		\textit{Constraints:} Max velocity 1 m/s, accel/decel $\pm$1 m/s$^2$, freq. 10 Hz, turning radius, \mfb{$E_l \geq E_l^{\rm min}$};
		\textit{Actions:} Move to position, align \\
		\hline
		Gripper & \textit{Resources:} Actuator, lifter, energy $E_g$;
		\textit{Constraints:} Max force, $m_{\rm max}$, geometry limits, freq. 10 Hz, \mfb{$E_g\geq E_g^{\rm min}$}; 
		\textit{Actions:} Grasp, release box \\
		\hline
		Mast & \textit{Resources:} Pan/tilt motors, energy $E_m$;
		\textit{Constraints:} Angular limits, freq. 10 Hz, \mfb{$E_m\geq E_m^{\rm min}$};
		\textit{Actions:} Look at given position \\
		\hline
		Camera & \textit{Resources:} Stereo camera (mast), mono camera (chassis), energy $E_c$;
		\textit{Constraints:} FOV, resolution, freq. 10 Hz, \mfb{$E_c \geq E_c^{\rm min}$};
		\textit{Actions:} Capture image \\
		\hline
	\end{tabular}
	\label{tab:usecase_subsystems}
\end{table}

\paragrafSimple{B. Petri Net Representation}

\begin{figure}
	\centering
	\includegraphics[width=1.0\linewidth]{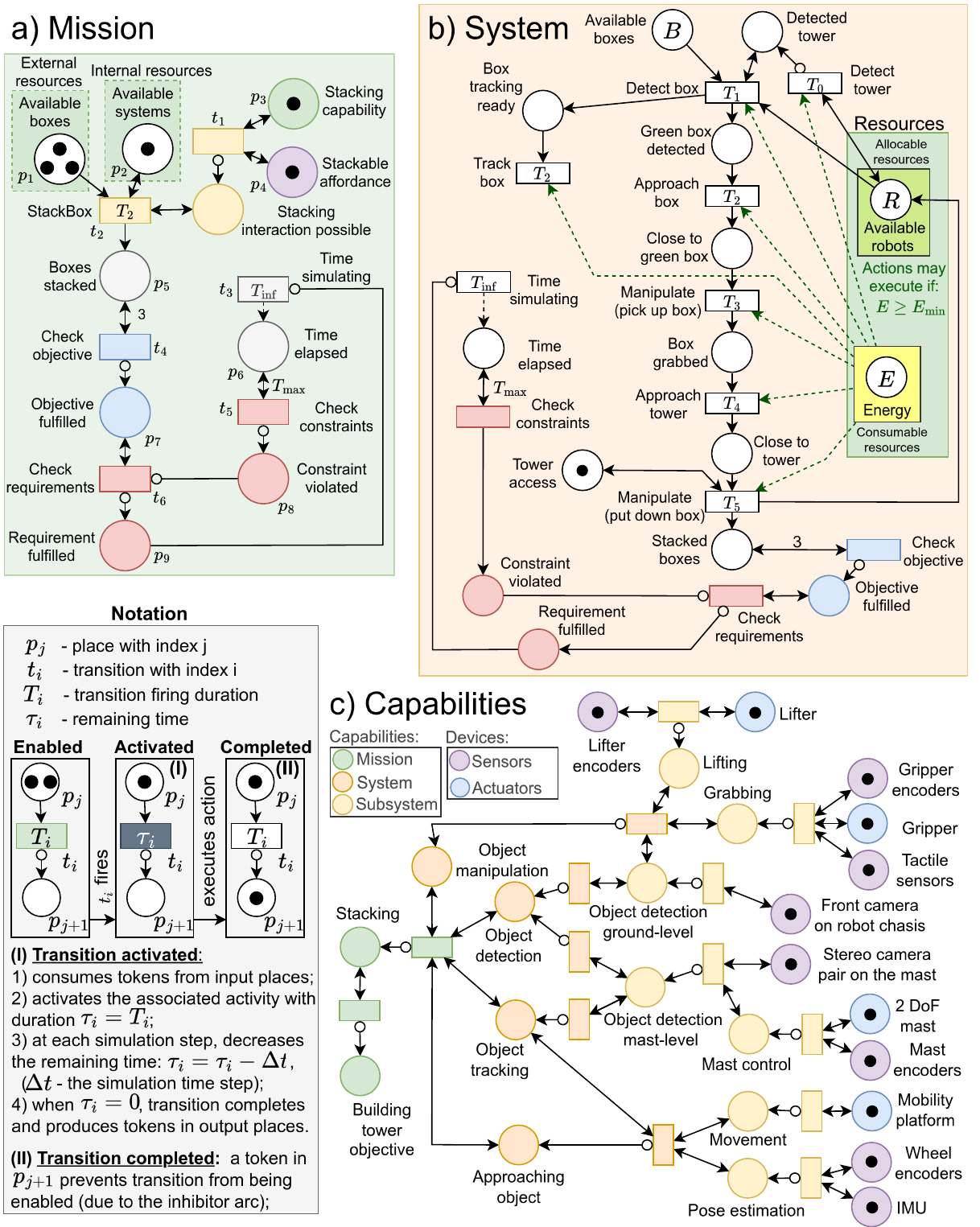}	
	\caption{PNs show: (a-b) mission- and system-level coordination, and (c) capabilities from three entity views. Green dashed arcs mark resources consumed/produced during execution.
	}
	\label{fig:usecase_mission_and_system}
\end{figure}

\noindent\mfb{\textbf{Mission Level:} Mission parameters are modeled using the extended PN in Fig.~\ref{fig:usecase_mission_and_system}a, capturing initial state, resources, and mission time constraints.  The \emph{Stack Box} action (transition $t_2$) has a defined duration \emph{$T_2$}. By tracking elapsed time, the PN verifies whether the task completes within limits.  Parallel \emph{Stack Box} actions, performed concurrently by multiple robots, may run by adjusting the initial marking to reflect available systems and resources, 
}

\noindent\mfb{\textbf{System Level:} The system-level PN (Fig.~\ref{fig:usecase_mission_and_system}b) decomposes the mission-level action into lower-level actions based on capabilities from integrated subsystems (e.g., sensors, actuators).  Action feasibility depends on the availability of functional subsystems and consumable resources (e.g., energy).  The PN supports evaluation of whether the objective can be achieved under current constraints.}

\noindent\mfb{\textbf{Subsystem Level:} Subsystem-level PNs capture individual capabilities, resources, and actions. Due to space constraints, they are omitted here but their simulation is shown in the accompanying video.}

\noindent{{\bf Rolling Up Combined Capabilities}:  \mfb{Fig.~\ref{fig:usecase_mission_and_system}c shows capabilities from all levels merged into a single PN, enabling integrated analysis across mission, system, and subsystem levels.  This capability PN allows evaluation of mission objective feasibility within allocated resources, including impact of subsystem failures, performed by removing tokens corresponding to disabled capabilities.}}

\paragrafSimple{C. PN Simulation and Analysis}

\mfb{To evaluate the feasibility of the robotic system design under real-world constraints, we simulate PNs capturing the Mission Level, the System Level, and a detailed view of Capabilities.  Included are task objectives, temporal constraints, resource limits, and failure probabilities.  The goal is to answer two questions:  (Q1) How many robots are required to reliably complete the task within resource limits, and (Q2) how does device reliability affect mission-level availability.}

\noindent\textbf{Simulation Setup:}
\mfb{Our PN simulator accepts single-run and Monte Carlo simulations using parameterized PN definitions.  For each transition, the user specifies an execution time distribution (e.g., normal or uniform), mean and variance, resource production / consumption rates, and weights for input / output arcs.  The initial marking encodes resource availability, entity state, and external environment state. Monte Carlo runs are performed by sampling these parameters from predefined ranges prior to each run.  Each PN is executed multiple times with randomized values for number of sub-entities (viz., robots), available energy, action durations, device reliability, and resource consumption rates.  The simulation provides timelines, token flow animations, state trajectories, and correlation matrices quantifying sensitivity to parameter variation.}

\noindent\textbf{Mission-Level Analysis (Q1):}
\mfb{Details of the case study analysis are provided in the accompanying video presentation.  At the mission-level, the PN (Fig.~\ref{fig:usecase_mission_and_system}a) is simulated to analyze how the number of participating robots affects ability to achieve the tower completion objective within the time limit.  It correctly predicts longer execution times (54.6 simulation units) for the single robot case as a consequence of sequential stacking actions.  Adding a second and third robot reduce expected time to 38.1 and 25.1 units respectively.  Four or more robots provide no benefit as only three stacking cubes exist.  Viewed as individual trials, Monte Carlo results predict a 72\% success rate within the 60 units  limit for one robot, with additional robots raising the success rate to 93\%.}

\begin{figure}
	\begin{subfigure}{0.4\linewidth}
		\centering
		\raisebox{3mm}{\includegraphics[width=1.0\linewidth]{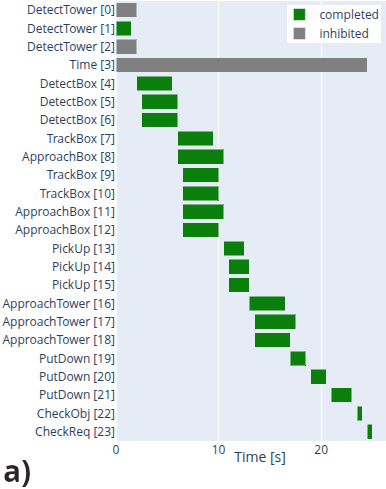}}		
	\end{subfigure}
	\begin{subfigure}{0.6\linewidth}
		\centering
		\includegraphics[width=1.0\linewidth]{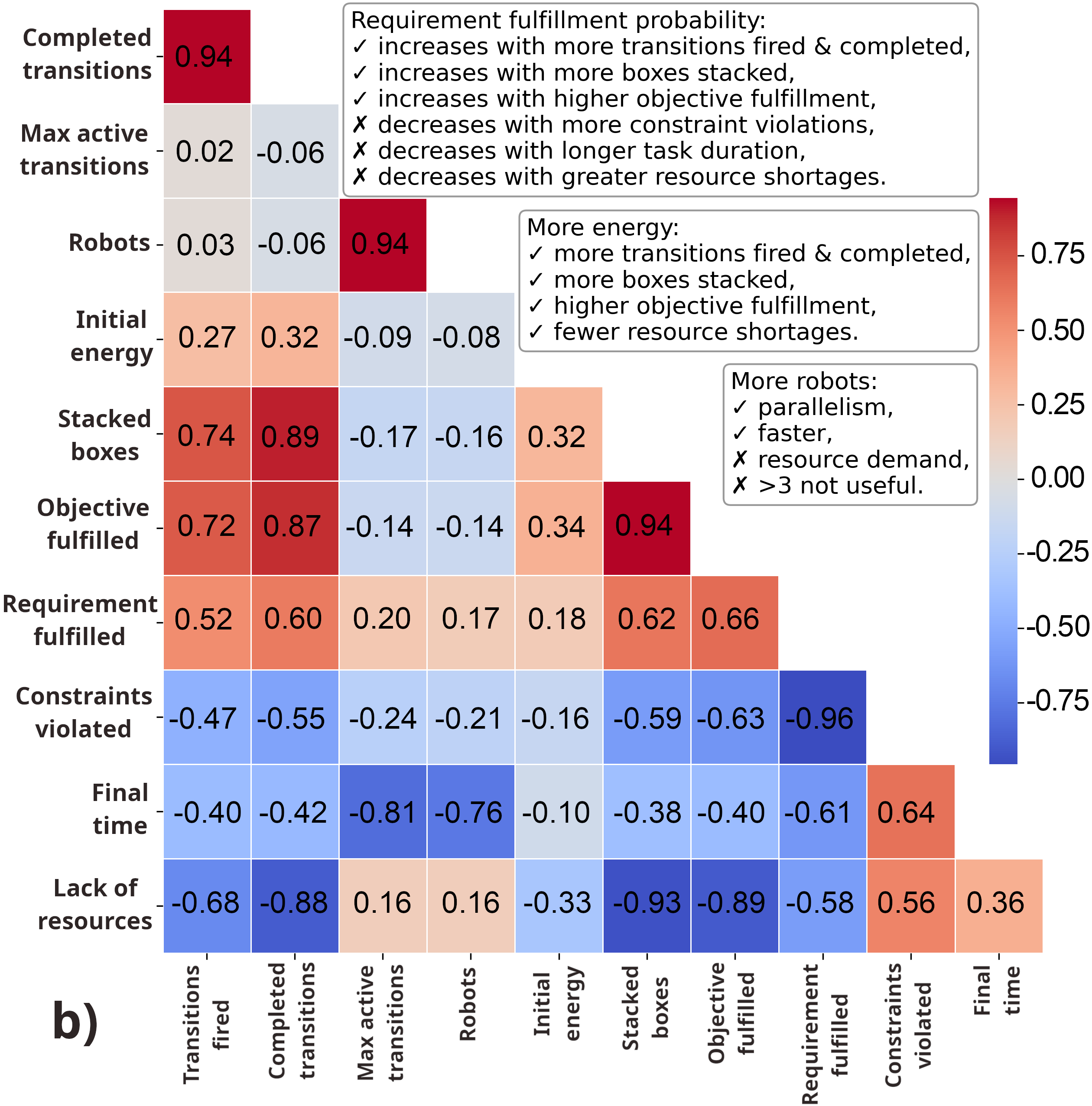}
	\end{subfigure}
		\caption{
		(a) Timeline for 3-robot system (PN in Fig.~\ref{fig:usecase_mission_and_system}b); green = success, gray = inhibited (1s sim = 10s real time); (b): Correlation matrix for system-phase PN (1000 runs).
		\mfb{Positive = direct correlation (e.g., $0.34$: more energy $\rightarrow$ higher probability of objective fulfillment); negative = inverse correlation (e.g., $-0.76$: more robots $\rightarrow$ shorter final time).}
		}
		\label{fig:usecase-sytem-both}
\end{figure}

\noindent\textbf{System-Level Analysis \mfb{(Q1)}:}
\mfb{The system-level PN model in Fig.~\ref{fig:usecase_mission_and_system}b models stacking execution via lower level capabilities, providing more insight into subsystem availability and resource constraints.  The correlation analysis (Fig.~\ref{fig:usecase-sytem-both}b) shows increasing number of robots concurrently active transitions reducing completion time (Fig.~\ref{fig:usecase-sytem-both}a), but at a cost in energy consumption.  This stems from interference between robots, viz., while one robot places a block, others must wait, with attendant increases in idle energy consumption.  The correlation matrix also confirms that higher initial energy improves the probability of reaching the objective within resource constraints. Such analyses reveal the balance between number of robots, resource allocation, and mission requirements.}

\noindent\textbf{Capability Analysis (Q2):}
\mfb{
Single execution of the Capability PN (Fig.~\ref{fig:usecase_mission_and_system}c) shows feasibility of mission objectives depending on device reliability and the capability hierarchy.  The PN links devices to subsystem, system, and mission-level capabilities, enabling closed-form availability estimates.  Device availability may be computed as $p_i^{\text{device}} = 1 - (1 - p_i)^{k_i}$, where $p_i$ is the reliability of an individual device and $k_i$ is its redundancy (assuming only one functional device is required).  Likewise, assuming independent devices and $N$ identical robots, $p_{\text{mission}} = 1 - \left(1 - \prod_i p_i^{\text{device}}\right)^N$, again assuming only a single robot must function to achieve the objective. 
}

\begin{figure}
	\begin{subfigure}{1.0\linewidth}
		\centering
		\includegraphics[width=1.0\linewidth]{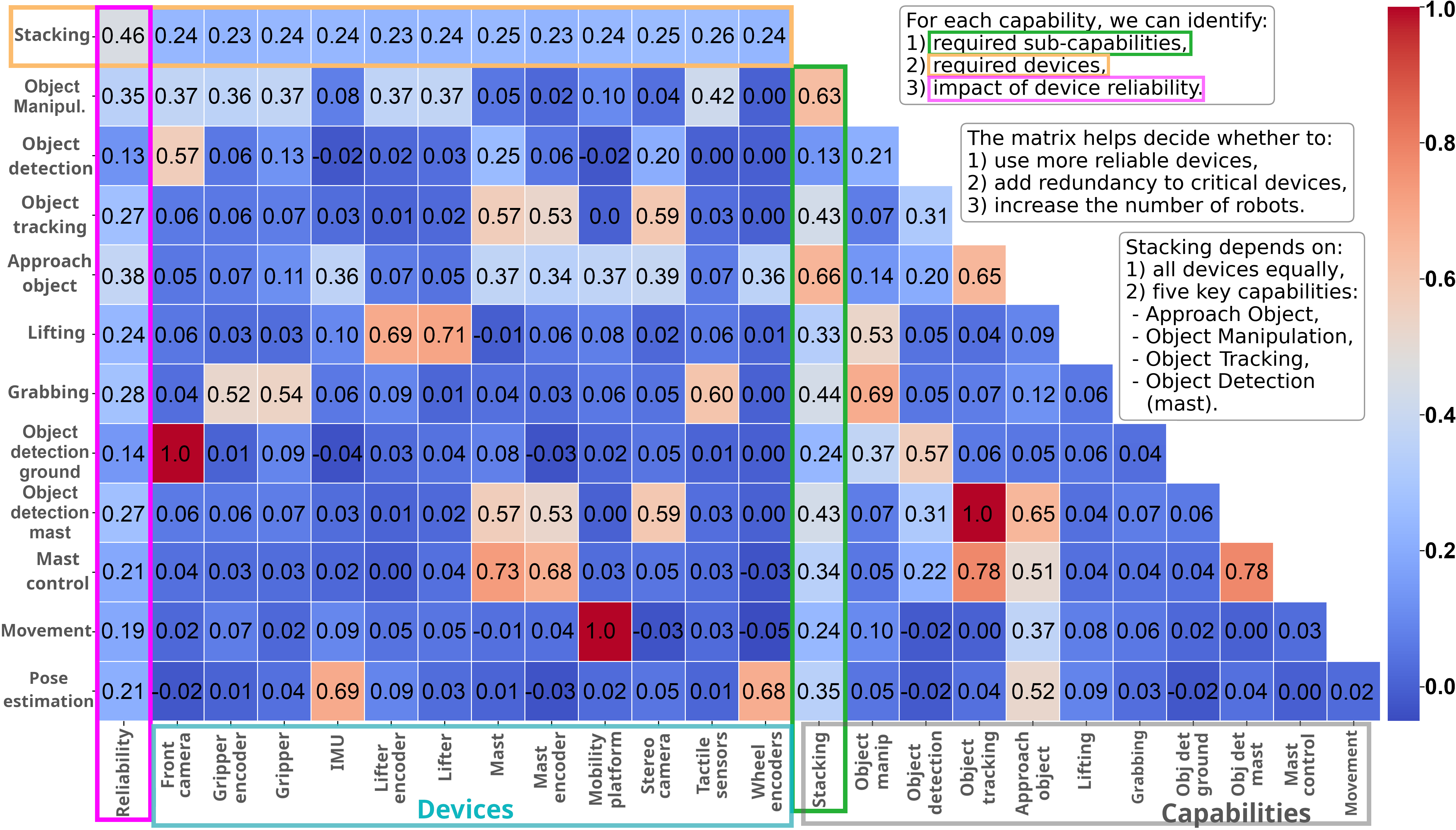}
	\end{subfigure}
	\begin{subfigure}{0.49\linewidth}
		\centering
		\includegraphics[width=1.0\linewidth]{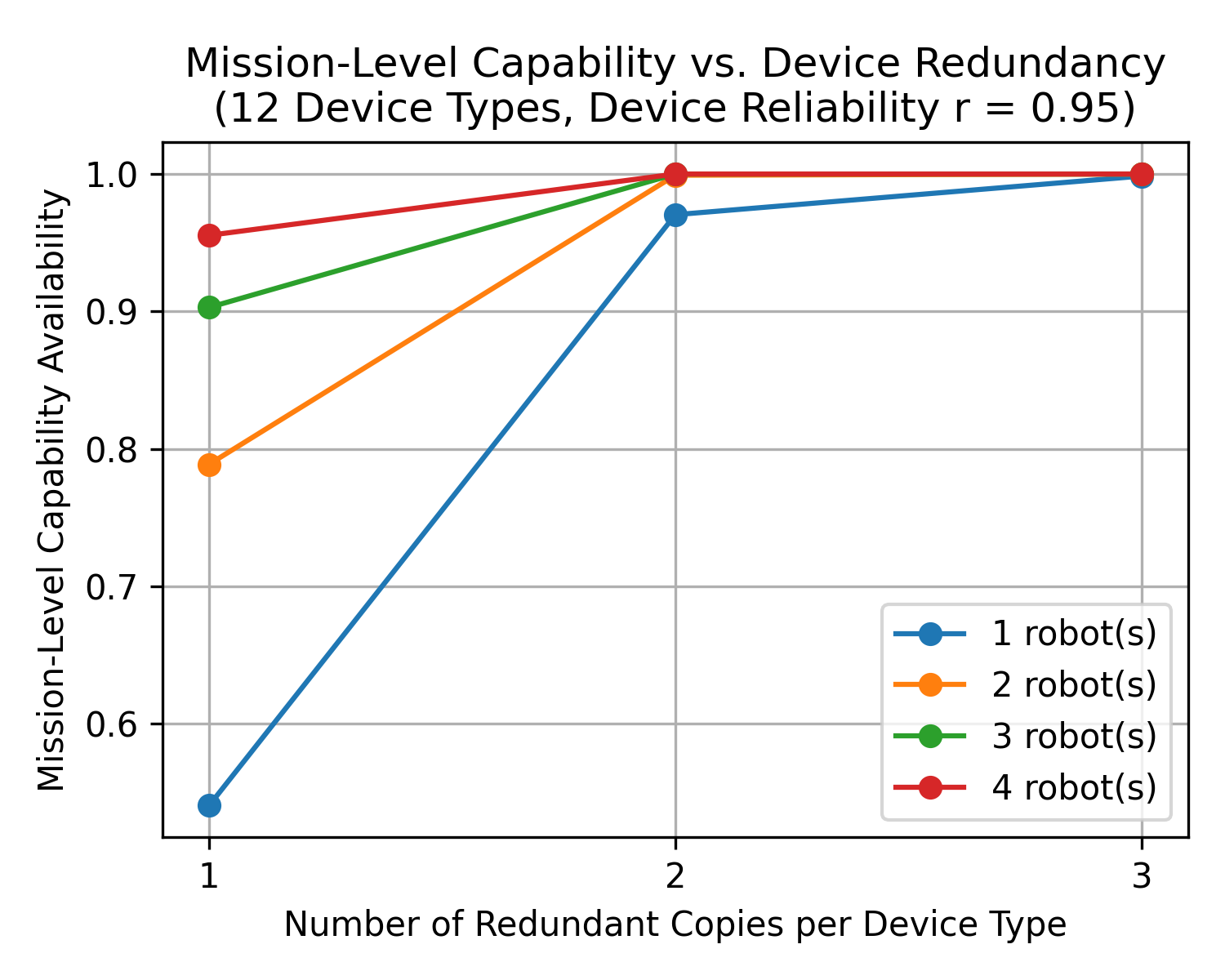}
	\end{subfigure}
	\begin{subfigure}{0.49\linewidth}
		\centering
		\includegraphics[width=1.0\linewidth]{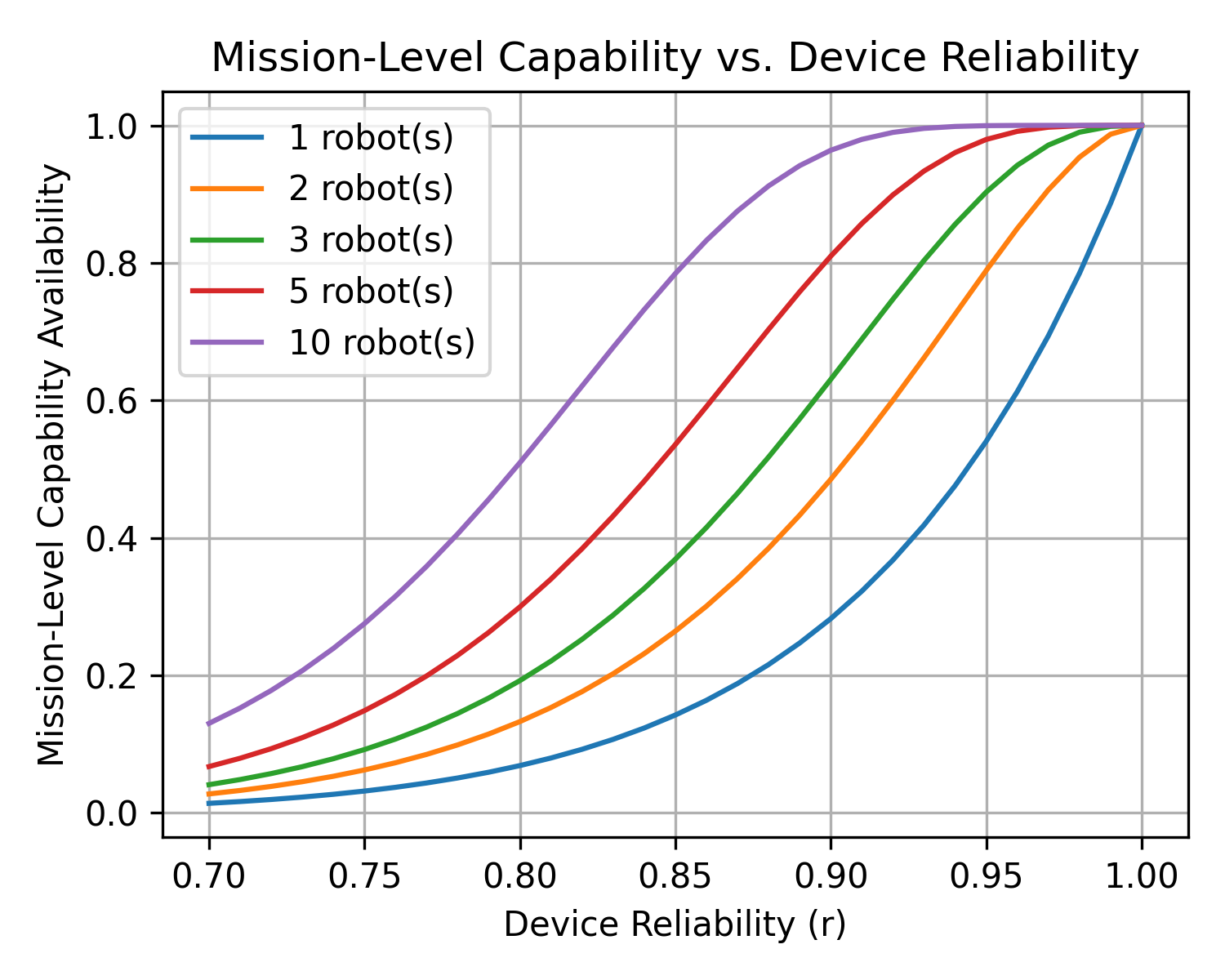}
	\end{subfigure}
	\caption{\textbf{Top:} Correlation matrix from capability PN (Fig.~\ref{fig:usecase_mission_and_system}c); 3 phases, 1000 runs, device reliability $\sim \mathcal{N}(0.9, 0.05)$.	
	\mfb{Left rectangle: capability-device correlations; right triangle: correlations among capabilities. Higher values indicate stronger relation (e.g. manipulation requires lifting, grabbing, detection).}
	\textbf{Bottom:} Mission-level capability availability: (left) subsystem redundancy ($r = 0.95$), (right) system redundancy.}
	\label{fig:experiments-reliability-redundancy-both}
\end{figure}

In complex systems, such dependencies are harder to trace, so we apply Monte Carlo simulations by randomizing device reliability, evaluating impact on capability and objective feasibility. The correlation matrix (top of Fig.~\ref{fig:experiments-reliability-redundancy-both}) identifies critical capabilities -- \textit{Approach}, \textit{Manipulation}, and \textit{Tracking}. The first row shows that mission-level capability depends equally on all devices -- failure of any disables it. The first column shows strongest impact of device reliability on mission-level \mfb{capability}. \mfb{Modifying PN structure (i.e., the connections between capabilities across adjacent layers) and adjusting redundancy via tokens in device-availability places enables exploration of alternative configurations and more informed architectural decisions.} \mfb{When adding robots (system-level redundancy) is infeasible, duplicating devices (subsystem-level redundancy) improves availability (see Fig.~\ref{fig:experiments-reliability-redundancy-both}, bottom: subsystem redundancy on the left, system redundancy on the right). Excessive subsystem-level redundancy may lead to cross-coupling, where interdependent devices increase complexity and risk of fault propagation via shared interfaces~\cite{Birolini:2017}.} This trade-off appears in NASA's CADRE~\cite{cadre:2024} mission, which uses three simple, homogeneous robots and a base station, rather than increasing subsystem-level redundancy.

\begin{figure}
	\centering
	\includegraphics[width=1.0\linewidth]{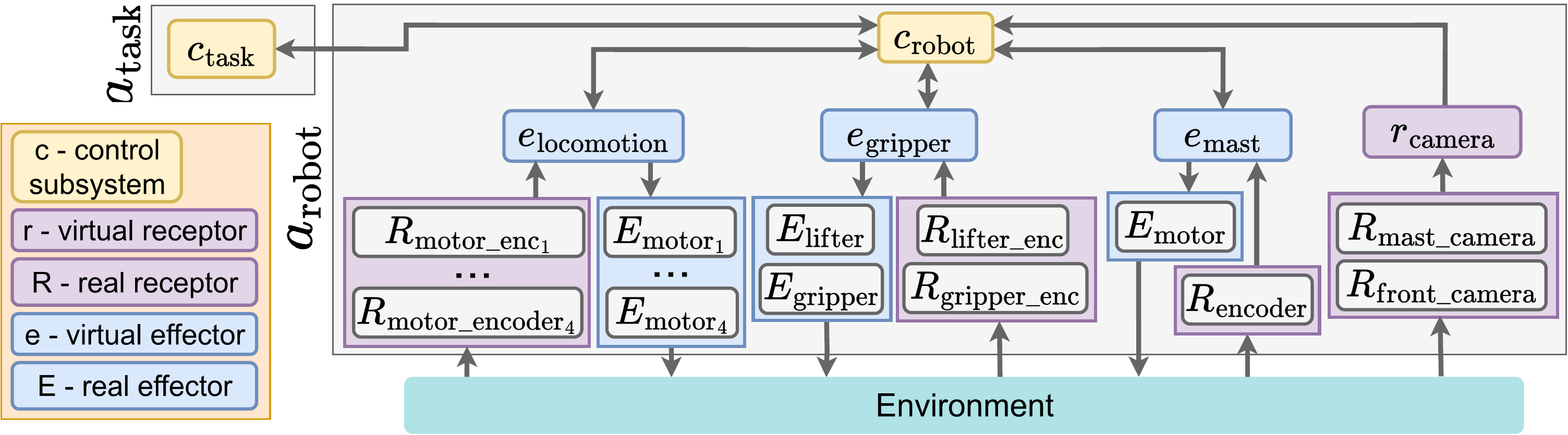}
	\caption{Robotic system structure for a single robot}
	\label{fig:use-case-rssm-structure}
\end{figure}

\vspace{1.0ex}
\noindent{\textit{D. Generated RSSM Parameters}}\\
Based on the 3S parameters, the RSSM parameters were derived, defining the robotic system's structure (Fig.~\ref{fig:use-case-rssm-structure}) and behavior (single hierarchical PN; omitted for space, see accompanying video). The system comprises two agents: $a_{\rm task}$, modeling the system task, and $a_{\rm robot}$, representing a robot composed of integrated subsystems. Each includes a control subsystem ($c_{\rm task}$ and $c_{\rm robot}$) managing its behavior, with direct communication between them. The robot agent contains virtual and real subsystems providing required capabilities. The task agent $a_{\rm task}$ handles task planning and monitoring, sending commands (corresponding to required capabilities) to $a_{\rm robot}$, which coordinates internal subsystems. The control subsystem $c_{\rm robot}$ \mfb{translates high-level commands from $c_{\rm task}$ (resulting from task decomposition) into effector commands, using sensory data, and provides feedback for closed-loop control.} This separation -- $a_{\rm task}$ maintaining objectives and \mfb{plans}, $a_{\rm robot}$ executing them -- supports task reassignment and multi-robot coordination. In multi-robot scenarios, e.g., two robots building one tower, a mission agent is added to plan and coordinate shared access/interaction, while each robot retains its task agent. \mfb{If robots build separate towers, the mission agent assigns each task agent to build a specific tower.}

\section{Experiments}
\label{sec:experiments} 

\begin{figure}
	\centering
	\includegraphics[width=1.0\linewidth]{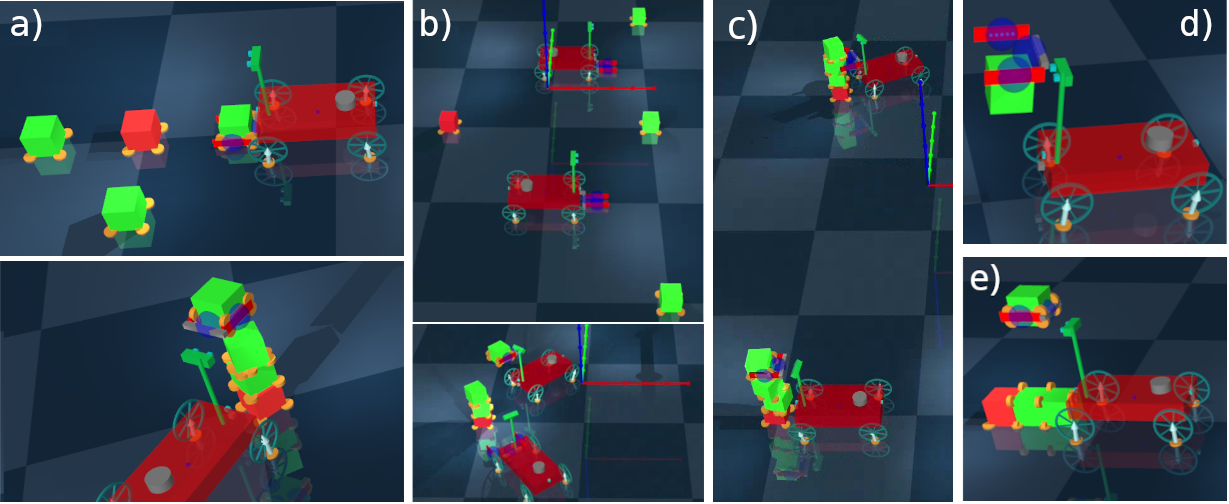}
	\caption{\textbf{Top:} Simulation views -- (a) 1 robot, 1 tower; (b) 2 robots, 1 tower; (c) 2 robots, 2 towers. 
		\textbf{Bottom:} Failures -- (d) unstable grasp; (e) collision/poor approach.
		}
	\label{fig:mujoco1}
\end{figure}

\noindent{\textbf{\underline{Simulation Setup}:}}
The RSSM parameters were expressed using the RSSL2 specification. This was achieved with a VSC plugin, developed as part of the \RSTM{} methodology providing parameterized RSSL2 templates. The RSSL2 compiler then automatically generated C++ code integrated with ROS~2. Each virtual and control subsystem was mapped to a separate ROS~2 node, communicating via topics. Physical effectors and sensors were modeled in MuJoCo~\cite{Todorov:2012} (Fig.~\ref{fig:mujoco1}), with a dedicated ROS~2 bridge -- also developed in this work -- handling data exchange. This standalone program published sensor data and enabled actuator control. In the simulation, the positions of the robot and blocks (green and red) were retrieved directly from MuJoCo.

\noindent{\textbf{\underline{Simulation Results}:}}
The system was evaluated across multiple scenarios to assess robustness and architectural flexibility. We varied: 1)~block dimensions and masses, 2)~subsystem execution frequencies, and 3)~task configurations with one or two robots building one or two towers.

\noindent\textit{Block dimensions:}
The robot operated reliably with blocks sized between 10-15\,cm. Smaller blocks tended to slip from the gripper, while oversized ones could not be lifted. In one test, a 4.5\,m block required collaboration between two robots, demonstrating the importance of modeling shared capabilities.

\noindent\textit{Execution frequencies:}
Subsystem frequencies influenced task performance. Low-frequency control loops introduced delays in locomotion, mast tracking, and grasping, resulting in jerky motion and missed interactions. Higher frequencies improved responsiveness but increased computational load. Table~\ref{tab:summary-frequencies-response} shows: raising subsystem frequency from 10\,Hz to 50\,Hz shortened task time by 11\,s,
\mfb{but at the cost of higher computational load.}
Lower worst-case control latencies improved failure response.

\noindent\textit{Task configurations:}
In the collaborative tower scenario with two robots (Fig.~\ref{fig:mujoco1}b), a single, complex task agent coordinated robots, adapting its logic to manage shared objective and resources. 
\mfb{In the dual-tower setup (Fig.~\ref{fig:mujoco1}c), the mission agent assigned resources (i.e., blocks) to task agents, so that each task agent together with its robot agent was responsible for building one tower.}
These tests show how task delegation and resource distribution influence system architecture and adaptability.

\noindent\textit{Affordance-related failures:}
Several unexpected behaviors were observed (Fig.~\ref{fig:mujoco1}d-e), including slipping blocks, collisions,
\mfb{failed lifts (size-dependent)}, and unstable placements. These issues stemmed not from system faults, but from missing \textit{negative affordances} -- concepts signaling risk or failure potential, such as slip risk, obstacle \mfb{risk}, and drop risk. Considering such affordances during design helps identify needed capabilities and guides architectural decisions, improving robotic system robustness and reliability.


\begin{table}
	\centering
	\caption{Test results with estimated worst-case response times for control loops (RT in ms), where: RT1 - motion control; RT2 - mast control; RT3 - gripper control; RT4 - task-agent communication.}
	\label{tab:summary-frequencies-response}
	\renewcommand{\arraystretch}{1.1}
	\setlength{\tabcolsep}{2.5pt}
	\begin{tabular}{|c|c|c|c|c|c|c|c|c|}
		\hline
		\textbf{Test} & \textbf{$f_{c_{\rm task}}$} & \textbf{$f_{\rm other}$} & \textbf{Time} & \textbf{Status} & \textbf{RT1} & \textbf{RT2} & \textbf{RT3} & \textbf{RT4} \\ \hline
		1 & 5Hz & 10Hz & 309s & Success & 400 & 400 & 300 & 400 \\ \hline
		2 & 5Hz & 50Hz & 298s & Success & 80 & 80 & 60 & 240 \\ \hline
		3 & 1Hz & 10Hz & 472s & Success & 400 & 400 & 300 & 1200 \\ \hline
		4 & 5Hz & 5Hz & -- & Failed & 800 & 800 & 600 & 600 \\ \hline
	\end{tabular}
\end{table}


\section{Conclusions}
\label{sec:conclusions}
This paper introduced \RSTM{}, a methodology that bridges the specification gap and supports architectural decision-making by transforming high-level robotic objectives into formal specifications at the mission, system, and subsystem levels. \RSTM{} combines ontological modeling with stochastic, timed, and resource-aware Petri nets for early-phase analysis of feasibility, timing, resource usage, and capability availability. As shown in Table~\ref{tab:method-comparison}, it differs from traditional MBSE approaches, which require
\mfb{a system model} without guiding how to develop it or what design decisions to make. In contrast, \RSTM{} leads the designer through iterative decision-making with intermediate, executable PN-based models \mfb{(e.g., analyzing how the number of robots or the available initial energy affect mission success and completion time)}, enabling key architectural properties to be verified before committing to a final system model and code generation. Recent Large Language Model (LLM)-based approaches~\cite{Jin:2024:RobotGPT,Hu:llm:2024} also operate in early phases but lack an internal, explainable model, making it difficult to analyze design rationale and limiting their role to action scheduling rather than producing formally grounded specifications.

A tower-building task simulation validated \RSTM{}, showing how ontological concepts guide development of executable specifications and inform architectural decisions. Experiments in configurations (e.g., one or two robots, shared or separate objectives) assessed how team size and coordination affect performance. \mfb{This method enables exploration of the design decision space, such as whether tasks are executed by a single robot or distributed among multiple robots, and whether objectives are pursued jointly or independently. While our current validation focuses on task distribution and resource allocation, the same framework can in principle be extended to support higher-level architectural choices (e.g., reactive vs. layered autonomy, centralized vs. decentralized multi-robot systems). Monte Carlo simulations further evaluate robustness under uncertainty, showing that two robots were sufficient to complete the task within time and energy limits, while device reliability strongly affected mission-level capabilities and objective fulfillment. Observed failure modes highlighted the importance of accounting for negative affordances to ensure robustness and recovery.}

Grounding \RSTM{} in ontological concepts also enables integration with LLMs~\cite{Jin:2024:RobotGPT,Hu:llm:2024} as interactive assistants in parameter derivation, supporting explainable, AI-assisted specification development. While this work demonstrated integration with RSSL2 for ROS 2-based controller synthesis, the approach is not limited to this toolchain and could also connect to frameworks such as NASA’s F Prime~\cite{Bocchino:2018:FPrime} for safety-critical missions. Future work includes applying \RSTM{} to the multi-robot CADRE mission~\cite{cadre:2024}, focusing on decentralized, resource-aware planning. In conclusion, \RSTM{} offers a systematic, explainable, and extensible pathway from high-level objectives to validated, executable specifications.

\begin{table}[t]
	\caption{Comparison of RS(TM)$^2$ with selected approaches.}
	\label{tab:method-comparison}
	\scriptsize
	\renewcommand{\arraystretch}{1.0}
	\setlength{\tabcolsep}{2pt}
	\resizebox{\columnwidth}{!}{%
		\begin{tabular}{|p{1.1cm}|c|p{1.5cm}|p{2.7cm}|c|c|c|p{0.85cm}|}
			\hline
			\textbf{Approach} & \textbf{Ont.} & \textbf{Input} & \textbf{Output / Use} & \textbf{Multi-lvl.} & \textbf{Exec.} & \textbf{Res.} & \textbf{Ref.} \\ \hline
			
			State-based & partial & Predefined formal model (PN/FSM) & Execution or post-spec.\ analysis; no model synthesis guidance & partial & partial & \xmark & \cite{Figat:2022:RAS,Pelletier:2025:RAS} \\ \hline
			LLM-based & \xmark & High-level task & Non-formal action sequence/code & \xmark & \xmark & \xmark & ~\cite{Hu:llm:2024,Jin:2024:RobotGPT} \\ \hline
			State Analysis & partial & Predefined state model & Verification/analysis; no model acquisition method & partial & partial & partial & \cite{Ingham:2005,Wagner:2012} \\ \hline
			SysML & partial & SysML diagrams & Documentation/blueprint; mappable to formal models & \cmark & \xmark & partial & \cite{Wagner:2012,Friedenthal:2015} \\ \hline
			RSSL & partial & Params for meta-model & Auto-generated RSHPN $\rightarrow$ ROS code & \cmark & partial & partial & 
			\cite{Figat:2022:RAL} \\ \hline
			\textbf{RS(TM)$^2$} & \cmark & High-level task/objective & Guided synthesis of validated multi-lvl.\ formal model $\rightarrow$ RSSL2 & \cmark & \cmark & \cmark & -- \\ \hline
			
		\end{tabular}%
	}
	\vspace{0.8mm}
	
	\scriptsize
	Ont. -- ontology with robotics-specific concepts; partial -- limited coverage.  
	Input -- formal model, parameters, or high-level task/objective.  
	Multi-lvl. -- supports multiple abstraction levels.  
	Exec. -- executable specification; in RS(TM)$^2$for early-phase specification analysis.
	Res. -- explicit handling of resources, including time.
\end{table}

\end{document}